\pdfoutput=1
\documentclass[letterpaper]{article} 
\usepackage{aaai2026}  
\usepackage{times}  
\usepackage{helvet}  
\usepackage{courier}  
\usepackage[hyphens]{url}  
\usepackage{graphicx} 
\usepackage{natbib}  
\usepackage{caption} 
\usepackage{pifont}
\usepackage{algorithm}
\usepackage{algorithmic}

\usepackage{newfloat}
\usepackage{listings}

\DeclareCaptionStyle{ruled}{labelfont=normalfont,labelsep=colon,strut=off} 
\floatstyle{ruled}
\newfloat{listing}{tb}{lst}{}
\floatname{listing}{Listing}
\usepackage{subcaption}
\usepackage{amsmath}
\usepackage{booktabs}
\usepackage{amsfonts}
\usepackage{multirow}

\usepackage{xcolor} 

\title{Soft Conflict-Resolution Decision Transformer for Offline Multi-Task Reinforcement Learning}

\author {
    Shudong Wang\textsuperscript{\rm 1 \rm 2 \rm 3}\equalcontrib, 
    Xinfei Wang\textsuperscript{\rm 1 \rm 2 \rm 3}\equalcontrib, 
    Chenhao Zhang\textsuperscript{\rm 1 \rm 2 \rm 3}\equalcontrib \thanks{Corresponding author.},
    Shanchen Pang\textsuperscript{\rm 1 \rm 2 \rm 3},
    Haiyuan Gui\textsuperscript{\rm 4},
    Wenhao Ji\textsuperscript{\rm 1 \rm 2 \rm 3},
    Xiaojian Liao \textsuperscript{\rm 5}
}
\affiliations {
    \textsuperscript{\rm 1}School of Computer Science and Technology, China University of Petroleum(East China)\\
    \textsuperscript{\rm 2}State Key Laboratory of Chemical Safety\\
    \textsuperscript{\rm 3}Shandong Key Laboratory of Intelligent Oil \& Gas Industrial Software\\
    \textsuperscript{\rm 4}School of Information and Control Engineering, Qingdao University of Technology\\
    \textsuperscript{\rm 5}State Key Laboratory of Software Development Environment, Beihang University, Beijing, China\\

    \{wangsd@, z24070040@s., zch@, pangsc@\}upc.edu.cn, 
}

\begin{document}

\maketitle

\begin{abstract}



Multi-task reinforcement learning (MTRL) seeks to learn a unified policy for diverse tasks, but often suffers from gradient conflicts across tasks. Existing masking-based methods attempt to mitigate such conflicts by assigning task-specific parameter masks. However, our empirical study shows that coarse-grained binary masks have the problem of over-suppressing key conflicting parameters, hindering knowledge sharing across tasks. Moreover, different tasks exhibit varying conflict levels, yet existing methods use a one-size-fits-all fixed sparsity strategy to keep training stability and performance, which proves inadequate. These limitations hinder the model’s generalization and learning efficiency.

To address these issues, we propose \textbf{SoCo-DT}, a \textit{Soft Conflict-resolution} method based by parameter importance. By leveraging Fisher information, mask values are dynamically adjusted to retain important parameters while suppressing conflicting ones. In addition, we introduce a Task-Aware Mask Update with Adaptive Sparsity strategy based on the Interquartile Range (IQR), which constructs task-specific thresholding schemes using the distribution of conflict and harmony scores during training. To enable adaptive sparsity evolution throughout training, we further incorporate an asymmetric cosine annealing schedule to continuously update the threshold. Experimental results on the Meta-World benchmark show that SoCo-DT outperforms the state-of-the-art method by 7.6\% on MT50 and by 10.5\% on the suboptimal dataset, demonstrating its effectiveness in mitigating gradient conflicts and improving overall multi-task performance.

\end{abstract}


\section{Introduction}

\noindent

\textbf{Offline Reinforcement Learning (Offline RL)}~\cite{levine2020offline} enables policy learning from static, pre-collected datasets without interacting with the environment, offering improved safety, lower data costs, and better deployability in domains such as robotics~\cite{kumar2021workflow}, autonomous driving~\cite{shi2021offline}, and healthcare~\cite{ghasemi2025personalized}. However, offline RL often lacks cross-task generalization and must relearn from scratch when faced with new tasks~\cite{teh2017distral}, limiting its scalability in complex environments. Multi-task Reinforcement Learning (MTRL) addresses this by jointly training on multiple tasks to improve generalization and efficiency~\cite{d2024sharing,lee2022multi,caruana1997multitask}. 
Combining the strengths of both, \textbf{Offline MTRL} leverages task-mixed datasets to enable knowledge transfer without environment interaction. Despite its promise, offline MTRL faces substantial practical challenges, as task and data heterogeneity often induce gradient conflicts that hinder effective parameter sharing~\cite{tang2023concrete,shi2023recon}.

Existing approaches for mitigating gradient conflict in MTRL can be broadly categorized into the following four directions: 
\ding{172} Model Architecture Methods Based on Parameter Sharing: These methods achieves multi-task learning by sharing parameters across tasks, typically through injecting task identifiers or task embeddings into the model to achieve task differentiation~\cite{xu2022prompting,he2023diffusion,he2025goal}.
\ding{173} Gradient Projection and Orthogonalization Methods: This class of methods minimizes conflicts by projecting, discarding, or rescaling conflicting gradient components~\cite{yu2020gradient,chen2020just,wang2020gradient,chai2022model}.
\ding{174} Optimal Module Routing Methods: These methods construct optimal module compositions for each task via routing or task-specific output heads~\cite{he2024not,yang2020multi,sun2022paco}.
\ding{175} Optimal Parameter Subspace Methods: These methods identify optimal parameter subspaces for each task via masking~\cite{hu2024harmodt,zhang2024learning,sun2020learning}.
Recent studies, such as HarmoDT~\cite{hu2024harmodt}, combine sequence modeling with task-specific masks to identify optimal parameter subspaces while retaining shared parameters, which significantly reduces inter-task gradient conflict. Using HarmoDT\cite{hu2024harmodt} and Prompt-DT\cite{xu2022prompting} as examples, this paper identifies two key issues in current approaches.

First, existing masking-based methods are too coarse-grained to effectively support knowledge sharing, limiting the model’s generalization and learning efficiency. For example, while HarmoDT alleviates gradient conflicts via binary masks, it ignores parameter importance and may inadvertently suppress parameters that are both important and conflicting, which in turn decrease the model performance by approximately 4.67\% (Motivation 1, \S \ref{subsec:paramshare}).

Second, the degree of conflict among different tasks varies, and a one-size-fits-all fixed sparsity will disrupt the optimal subspace of some tasks. For example, HarmoDT applies a uniform sparsity of 20\% to conflicting parameters to maintain overall training stability and performance. However, in tasks such as \textit{basketball} and \textit{door-lock}, where the actual conflict ratio reaches 40\%–45\%, this fixed sparsity fails to provide sufficient coverage. As a result, it may retain too many conflicting parameters or mistakenly suppress important ones, leading to a performance drop of up to 60\% on those tasks and an average degradation of about 5.33\% overall (Motivation 2, \S \ref{subsec:softmask}).

To address these issues, we propose a novel soft masking mechanism tailored for offline MTRL, which enables fine-grained and importance-aware conflict mitigation. Unlike traditional binary masking strategies that indiscriminately suppress all conflicting parameters, our method assigns each conflicting parameter a soft mask value proportional to its task-specific Fisher ~\cite{zhang2024learning}. This allows important yet conflicting parameters to be retained with high weights, while less relevant ones are gradually suppressed—striking a principled balance between conflict alleviation and knowledge preservation.

Moreover, we introduce TAMU (Task-Aware Mask Update with Adaptive Sparsity), a unified masking strategy designed to improve parameter sharing efficiency and training stability. At its core, TAMU computes fine-grained conflict and harmony scores for each parameter, capturing both gradient consistency and magnitude imbalance across tasks. 
Built upon these scores, TAMU integrates two key mechanisms:
(1) \textit{a magnitude-sensitive harmony scoring mechanism}, which suppresses pseudo-consistent gradients with extreme magnitude deviations using a ReLU-based gating function with a tunable tolerance factor.
(2) \textit{a task-adaptive sparsity control module}, which determines masking thresholds based on the Interquartile Range (IQR) of conflict scores and dynamically adjusts them using an asymmetric cosine annealing scheduler, enabling task-sensitive sparsity evolution throughout training.
Together, these mechanisms enable TAMU to adaptively evolve the mask during training, resulting in improved multi-task generalization and stable learning dynamics across diverse offline tasks.

Experimental results on Meta-World benchmark demonstrate that our method consistently outperforms existing baselines, achieving up to 10.50\% improvement in average success rate while preserving parameter sharing efficiency.

\textbf{In summary, our main contributions are as follows:}

\begin{itemize}
    \item We identify key limitations in conventional mask-based MTRL methods in handling gradient conflicts.
    \item We propose a soft mask mechanism and task-aware mask update with adaptive sparsity strategy for more reliable subspace construction.
    \item We validate our method on multiple task combinations within the Meta-World benchmark, demonstrating consistent outperformance over state-of-the-art methods.
\end{itemize}

\section{Preliminary}
\subsection{Decision Transformer and Prompt-DT}
Decision Transformer (DT)~\cite{chen2021decision} formulates reinforcement learning as a conditional sequence modeling problem. Rather than estimating value functions or policy gradients, it directly predicts actions using a Transformer architecture~\cite{vaswani2017attention}.
At each timestep, the model takes as input a trajectory sequence consisting of return-to-go values \( R_t = \sum_{t'=t}^T r_{t'} \), states \( s_t \), and actions \( a_t \), forming the input sequence
$\tau = (\widehat{R}_1, s_1, a_1, \widehat{R}_2, s_2, a_2, \dots, \widehat{R}_T, s_T, a_T)$,
and outputs the next action via autoregressive prediction through a causally masked Transformer. The model is trained with a standard behavior cloning loss:
$\mathcal{L}_{DT} = \mathbb{E}_{\tau \sim \mathcal{D}} \left[ \sum_{t=1}^T \left\| \pi_\theta(\tau_{1:t}) - a_t \right\|_2^2 \right]$.

To enable generalization across tasks, Prompt-DT~\cite{xu2022prompting} extends DT to the multi-task setting via a prompting mechanism. For each task \( T_i \), a task-specific prompt is constructed from a short \( K \)-step demonstration subtrajectory sampled from the target task which is prepended to the online trajectory and used as additional input to encode task identity:
\begin{align}
    \tau^{\text{input}}_{i, t} = (&\widehat{R}^*_{i, 1}, s^*_{i, 1}, a^*_{i, 1}, \dots, \widehat{R}^*_{i, K}, s^*_{i, K}, a^*_{i, K}, 
    \widehat{R}_{i, t-K+1},\nonumber \\ &s_{i, t-K+1}, a_{i, t-K+1}, \dots, \widehat{R}_{i, t}, s_{i, t}, a_{i, t}),
\end{align}
this design enables Prompt-DT to adapt to new tasks in a few-shot fashion, without requiring fine-tuning.

\subsection{HarmoDT}
\label{harmodt}
HarmoDT~\cite{hu2024harmodt} extends Prompt-DT by introducing a task-specific mask mechanism that enables efficient and conflict-aware parameter sharing. For each task \( T_i \), HarmoDT learns a binary mask vector \( M^{T_i} \in \{0,1\}^d \), which selects a task-specific subspace from the full parameter set \( \theta \in \mathbb{R}^d \). The masked parameter is defined as \( \theta^{T_i} = \theta \odot M^{T_i} \), where \( \odot \) denotes element-wise multiplication.

HarmoDT jointly optimizes the task-specific masks \( M = \{M^{T_1}, \dots, M^{T_N}\} \) and shared model parameters \( \theta \),aiming to maximize the expected cumulative return under fixed masks while minimizing the multi-task training loss:
\begin{align}
\max_{M}~ & \mathbb{E}_{T_i \sim p(T)}\left[ \sum_{t=0}^{\infty} \gamma^t \mathcal{R}^{T_i}(s_t,\, \pi(\tau_{i,t}^{\text{input}} \mid \theta^{*T_i})) \right], \\
\text{s.t.}~ & \theta^* = \arg\min_{\theta} \mathbb{E}_{T_i \sim p(T)} \mathcal{L}_{DT}(\theta, M), \\
\text{where}~ & \theta^{*T_i} = \theta^* \odot M^{T_i}, \quad M = \{M^{T_i} \}_{T_i \sim p(T)}.
\end{align}

To identify the optimal harmony subspace for each task, HarmoDT computes a harmony score for each parameter by combining the agreement score \( A(T_i) = \bar{g}_i \odot \frac{1}{N} \sum_{i=1}^N \bar{g}_i \), which reflects gradient alignment, and the importance score \( F(T_i) =\left(\nabla \log \mathcal{L}_{T_i}\left( \theta^{T_i}\right) \odot M^{T_i} \right)^2 \), derived from the Fisher information. The combined harmony score is defined as \( H(T_i) = A(T_i) + \lambda F(T_i) \), where \( \lambda \) balances agreement and importance.
HarmoDT maintains an equal sparsity level for each task by removing and recovering the same number of parameters during training, thereby enabling the discovery of an optimal subspace with
consistent sparsity. It achieves state-of-the-art performance on the Meta-World benchmark~\cite{yu2020meta} across environments with 5, 30, and 50 tasks (denoted as MT5, MT30, and MT50).

\begin{figure}[h]
  \centering
  \begin{subfigure}{0.23\textwidth}
   \centering
  \includegraphics[width=\linewidth]{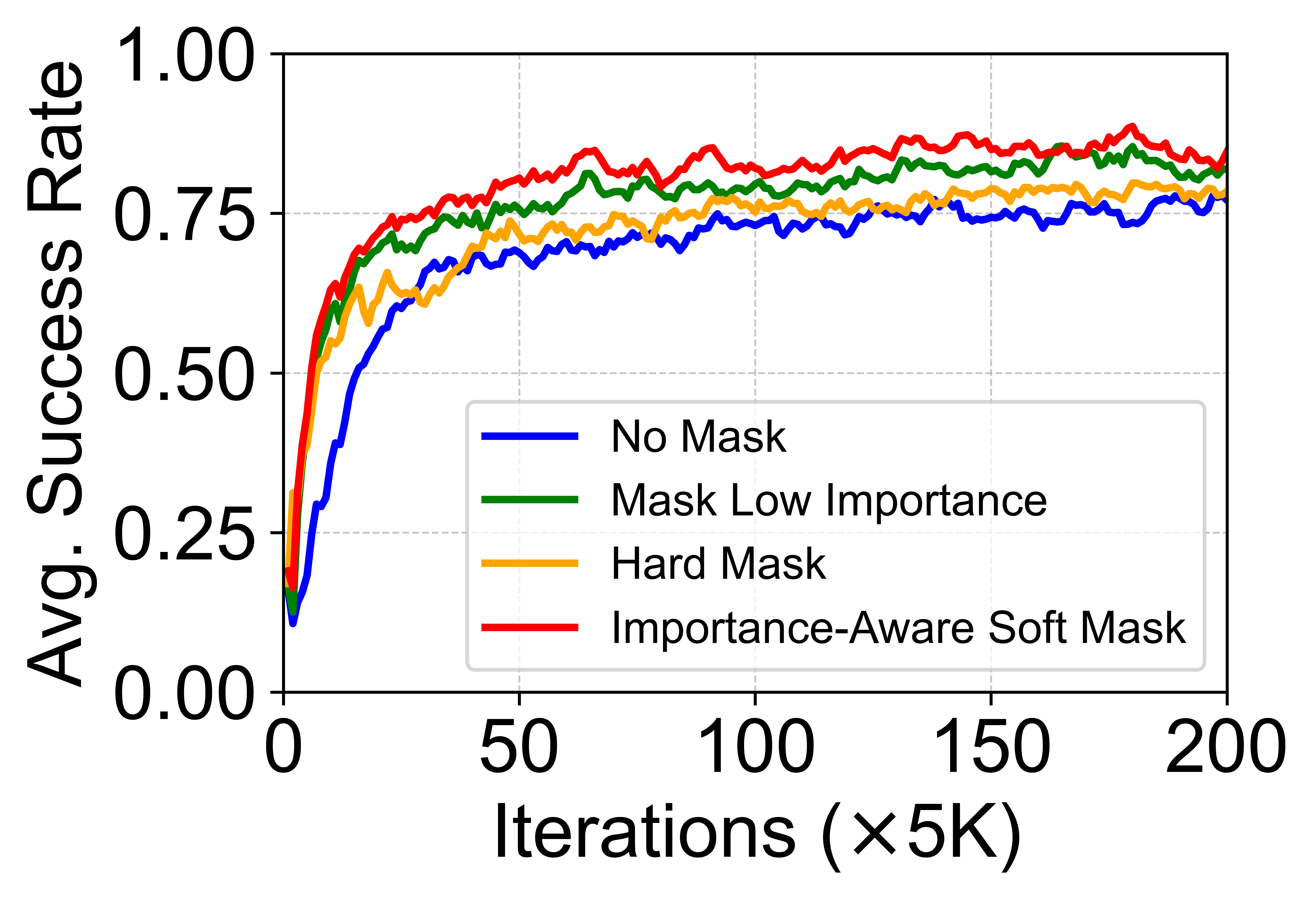}
  \caption{Near-optimal datasets}
  \label{fig:harmony_comparison}
  \end{subfigure}
  \hfill
  \begin{subfigure}{0.23\textwidth}
    \centering
    \includegraphics[width=\linewidth]{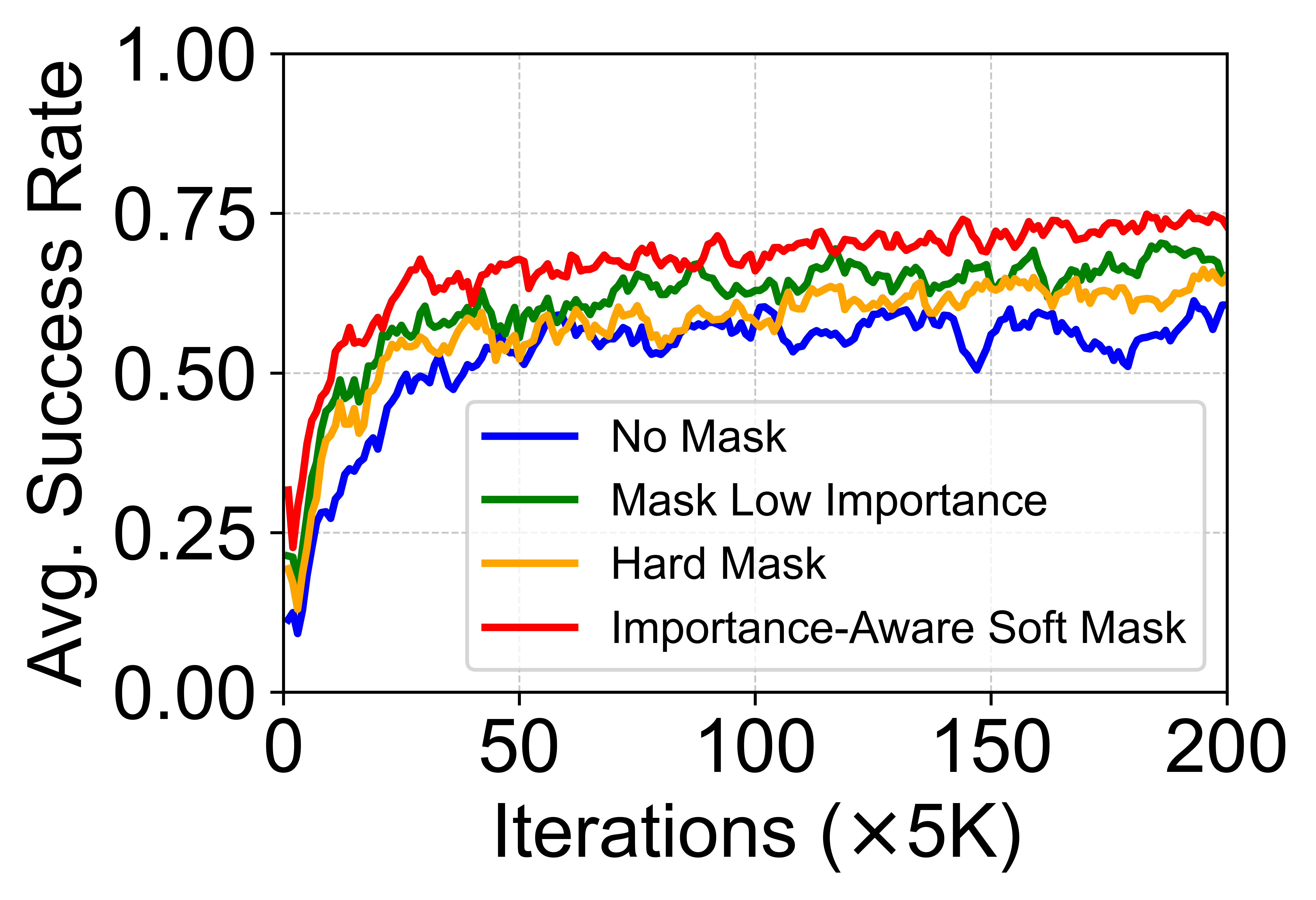}
    \caption{Sub-optimal datasets}
    \label{fig:harmony_comparison_sub}
  \end{subfigure}

\caption{Average success rate of four conflict-handling strategies on MT15:
\ding{172} PromptDT (no mask),
\ding{173} HarmoDT (hard mask, SOTA),
\ding{174} Ours (soft mask, ideal).
(a) Meta-World with near-optimal datasets;
(b) Meta-World with sub-optimal datasets. See §\ref{sec:experiments} for details.
}
\label{fig:masked_compare}
\end{figure}
\section{Rethinking the Mask in MTRL}\label{rething}

Through extensive exploration of existing mask-based works, we have two key motivations that demonstrate the vast potential for improvement in offline MTRL analysis, as shown in Fig. ~\ref{fig:masked_compare} and ~\ref{fig:masked_count}.

\subsection{Parameter-Sharing}\label{subsec:paramshare}
\textbf{Motivation \uppercase\expandafter{\romannumeral1}.} Coarse-grained binary masks make it difficult for tasks to share knowledge, limiting the generalization ability and learning efficiency of the model.

In MTRL, conflicting parameter updates across tasks are a key factor limiting generalization. Prompt-DT\cite{xu2022prompting} and HarmoDT\cite{hu2024harmodt} represent the two extremes of the conflict handling strategy: Prompt-DT does not mask the parameters of conflicts, while HarmoDT employs a binary mask to completely block conflicting parameters to alleviate inter-task interference—achieving a 6.8\% performance gain Fig.~\ref{fig:harmony_comparison}. However, our further observation that such hard masking approaches have increasingly evident limitations: they tend to unintentionally suppress parameters that are crucial for individual tasks. As shown in Fig.~\ref{fig:masked_count}(b), we find that approximately 1,550 top-importance parameters (within the top 30\%) are mistakenly masked due to binary conflict handling. When we manually restore these parameters, the average task success rate improves by 4.45\%. On sub-optimal datasets, the gain increases to 3.92\%, as illustrated by the green curve in Fig.~\ref{fig:masked_compare}. These results suggest that rigid masking sacrifices valuable knowledge and weakens parameter sharing, particularly when tasks diverge significantly or datasets contain many sub-optimal trajectories.

\textbf{Such substantial improvement highlights the urgent need for the design of more balanced and resilient masking strategies in MTRL.} We validate this hypothesis by adopting a soft masking mechanism (\S\ref{soft masking mechanism}). As shown by the red curve in Fig.\ref{fig:masked_compare}, the soft mask can effectively avoid unnecessary parameter suppression, achieving a notable 6.96\% performance gain. Moreover, under more challenging conditions involving sub-optimal datasets, this ideal mechanism achieves 9.33\% potential performance improvement.

\subsection{Multi-Task Training Stability}\label{subsec:softmask}
\textbf{Motivation \uppercase\expandafter{\romannumeral2}.} A fixed sparsity level is insufficient to accommodate the complexity and variability of diverse tasks.

\begin{figure}[h]
  \centering
  \begin{subfigure}{0.23\textwidth}
   \centering
  \includegraphics[width=\linewidth]{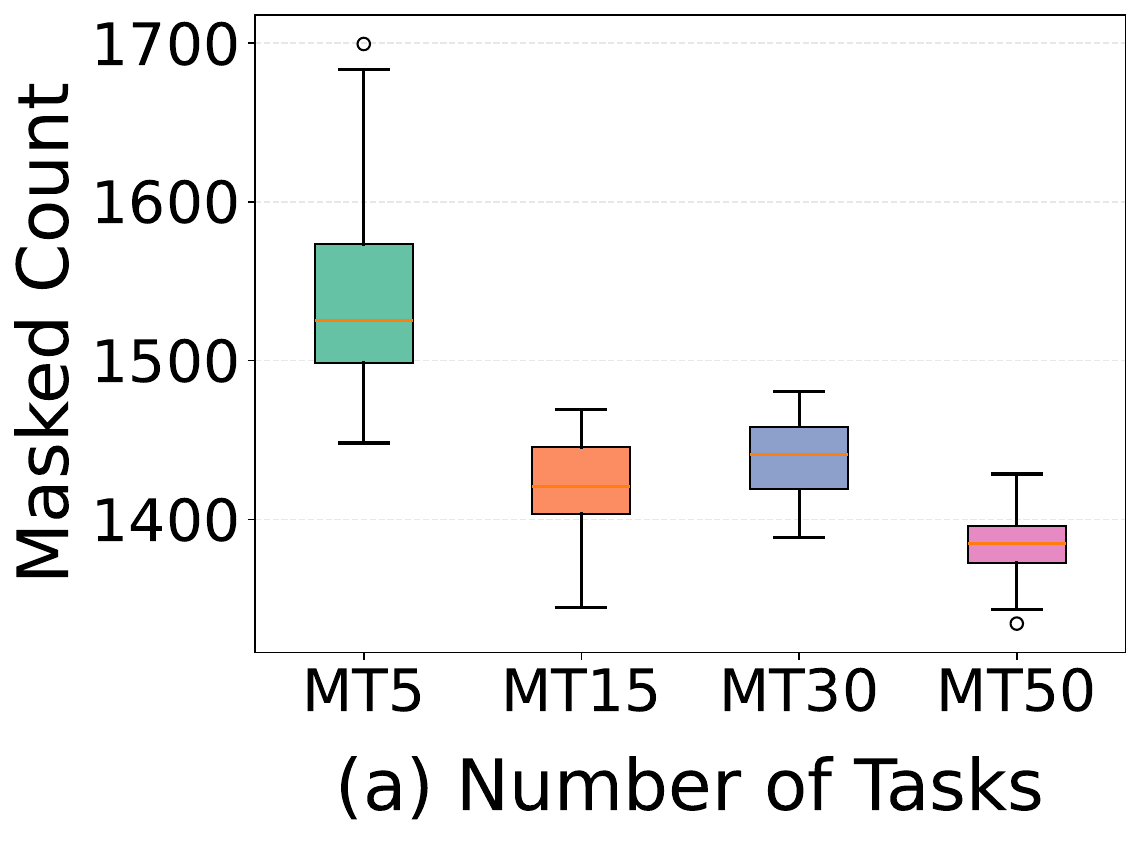}
  \label{fig:boxplot}
  \end{subfigure}
  \hfill
  \begin{subfigure}{0.23\textwidth}
    \centering
  \includegraphics[width=\linewidth]{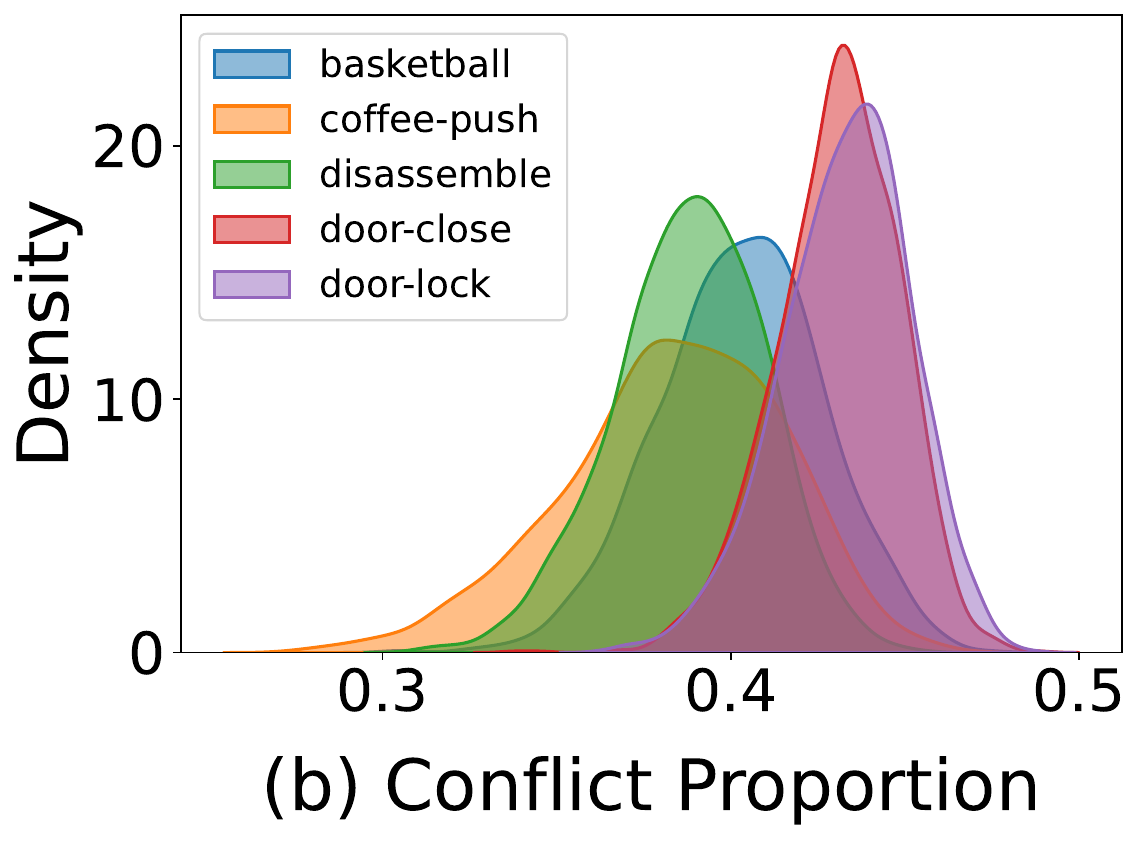}
  \label{fig:mot3}
  \end{subfigure}
\caption{
(a) Average number of important parameters wrongly masked during training under Fisher-based detection.
(b) Distribution of conflicting parameters among tasks.
}
  \label{fig:masked_count}
\end{figure}

As the number of tasks increases, inter-task interactions become more complex, resulting in significant variation in conflict intensity. To stabilize training, existing mask-based methods~\cite{hu2024harmodt,zhang2024learning,sun2020learning} typically impose a fixed sparsity level on the mask matrix, aiming to limit the number of active parameters and prevent unstable updates. However, this static strategy proves inadequate in heterogeneous task scenarios, where conflict patterns vary considerably both across tasks and over time. Fig.~\ref{fig:masked_count}(a) illustrates the distribution of conflicting parameters at convergence under the MT15 setting using Prompt-DT without masking. The five selected tasks differ markedly in their target distributions. The results reveal two key findings: \textbf{(1) the proportion of conflicting parameters differs significantly across tasks, and (2) the conflict ratios exhibit dynamic changes over the course of training}. Obviously, a one-size-fits-all fixed sparsity constraint struggles to accommodate task-specific conflict structures, making it difficult to discover an optimal shared parameter subspace. \textbf{This highlights the need for a strategy that balances training stability with adaptability to task-specific characteristics.}

\section{Method}
\begin{figure*}[h]
    \centering
    \includegraphics[width=0.85\textwidth]{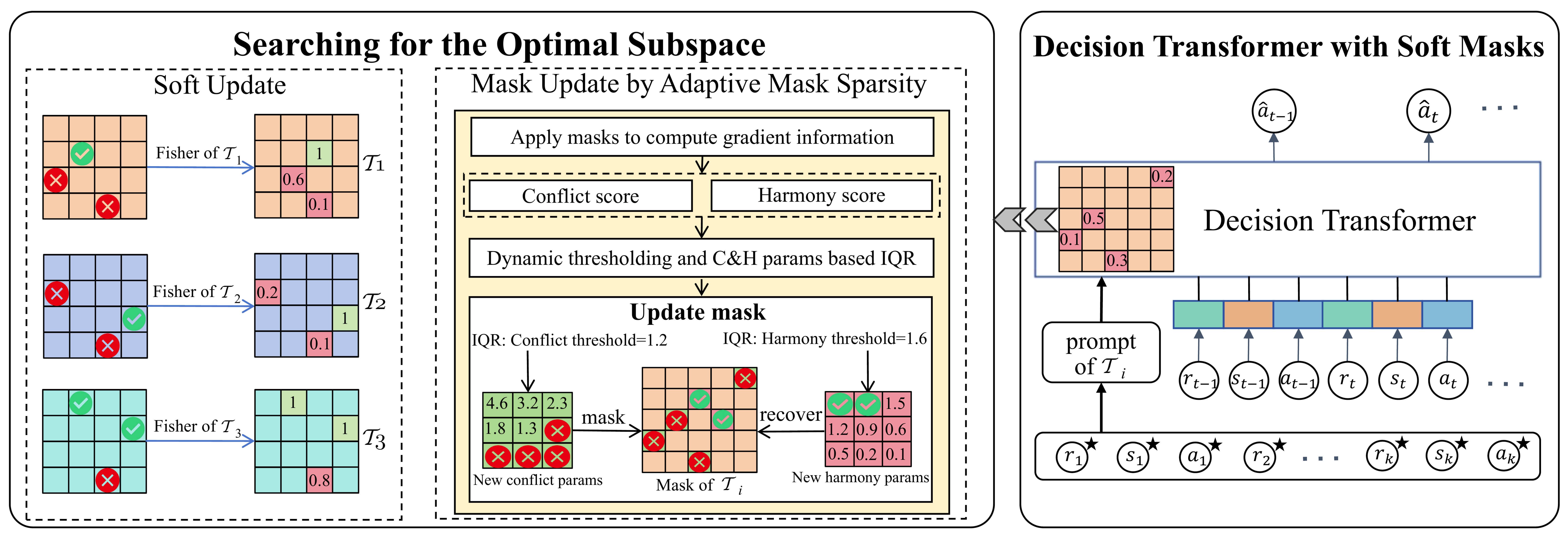}
    \caption{Illustrates the overall framework of SoCo-DT. The left panel shows the process of identifying task-optimal subspaces using the soft masking strategy and TAMU. The right panel presents the workflow of SoCo-DT based on a prompt-enhanced Decision Transformer architecture.}
    \label{fig:method_figure}
\end{figure*}

Our SoCo-DT framework consists of three key components: (i) an Importance-Aware Soft Masking Mechanism (§\ref{soft masking mechanism}), (ii) a Task-Aware Mask Update with Adaptive Sparsity Strategy (§\ref{subsec:Task-Adap}), The overall framework and training pipeline are shown in Figure~\ref{fig:method_figure} and Algorithm~\ref{alg:erk-finegrained-mask}.

\subsection{Importance-Aware Soft Masking Mechanism}
\label{soft masking mechanism}

To balance conflict suppression and parameter preservation, we assign a soft importance weight to each conflicting parameter based on its significance to the current task. Specifically, for each task $T_i$ ($i = 1, \dots, N$) with loss $\mathcal{L}_{T_i}$, parameter subspace $\theta^{T_i}$, and mask matrix $M^{T_i}$, the Fisher information~\cite{hu2024harmodt} is estimated as:

\begin{equation}
\label{fisher}
F(T_i) = \left( \nabla \log \mathcal{L}_{T_i}(\theta^{T_i}) \odot M^{T_i} \right)^2,
\end{equation}
where $\odot$ represents element-wise multiplication. We normalize Fisher information across all parameters of task ${T_i}$. The soft mask for parameter 
$j$ is computed as:

\begin{equation}
M^{T_i}_j =
\begin{cases}
1 & \text{if parameter is in harmony,} \\[8pt]
\displaystyle \frac{F(T_i)_j - F_{\min}}{F_{\max} - F_{\min}} & \text{if parameter is conflicting,}
\end{cases}
\end{equation}
where $F_{\min} = \min_j F(T_i)_j, F_{\max} = \max_j F(T_i)_j$.

During the forward pass, we apply a binary mask to retain task-specific parameters. During the backward pass, we perform discounted gradient correction on conflicting parameters, allowing limited yet informed updates without disrupting prior tasks.
\begin{equation}
\theta_{t+1} = \theta_t - \eta \, \mathbb{E}\left[ \nabla \mathcal{L}_{\mathcal{T}_i}(\theta \odot \widetilde{M}^{T_i}) \right] \odot M^{T_i},
\end{equation}
where $\widetilde{M}^{T_i} = \mathbb{I}(M^{T_i} = 1)$ is the binary mask applied during the forward pass, $M^{T_i} \in [0,1]$ is the soft mask used to control the magnitude of gradient updates for conflicting parameters, and $\eta$ denotes the learning rate.

\subsection{TAMU: Task-Aware Mask Update with Adaptive Sparsity}
\label{subsec:Task-Adap}

During the mask update phase, we propose TAMU, a task-aware strategy with adaptive sparsity that enhances parameter sharing while ensuring training stability. It estimates each parameter’s conflict and harmony scores via a magnitude-sensitive metric, then applies task-specific thresholds to guide masking. Each task’s mask is updated through a unified strategy.

\textbf{Conflict Score.} For each task $T_i$ and parameter $j$, we compute the element-wise product between the task gradient $\mathbf{g}_i$ and the average gradient $\bar{\mathbf{g}}_j$, and incorporate the Fisher information $F(T_i)_j$  to reflect parameter importance. The conflict score is defined as:

\begin{equation}
\label{Cconflict}
C_{\text{conflict}}(T_i)_j = (\mathbf{g}_i \odot \bar{\mathbf{g}})_j + \lambda F(T_i)_j,
\end{equation}
where $\lambda$ is a trade-off coefficient that balances gradient alignment and parameter importance. A lower score indicates stronger conflict, and thus the parameter is more likely to be masked.

\textbf{Harmony Score.} For each task $T_i$, we refine the dot-product term $(\mathbf{g}_i \odot \bar{\mathbf{g}})$ using a magnitude-sensitive modulation factor $H(T_i)$, designed to downweight parameters with large magnitude deviations. Specifically, we define $H(T_i)$ as a ReLU-based gating function:

\begin{equation}
\label{Hti}
H(T_i) =\mathrm{ReLU} \left( \frac{ \alpha |\bar{\mathbf{g}}| - |\mathbf{g}_{i}| }{ \alpha|\bar{\mathbf{g}}|} \right),
\end{equation}
$\alpha$ is a tunable hyperparameter that controls the acceptable tolerance range. The Harmony score is then defined as:

\begin{equation}
\label{Charmony}
C_{\text{harmony}}(T_i)_j =
\begin{cases}
(\mathbf{g}_i \odot \bar{\mathbf{g}})_j \cdot H(T_i)_j  & (\mathbf{g}_i \odot \bar{\mathbf{g}})_j > 0, \\
(\mathbf{g}_i \odot \bar{\mathbf{g}})_j, & \text{otherwise}.
\end{cases}
\end{equation}
By suppressing parameters with large inter-task gradient magnitude discrepancies, this method prevents individual tasks from dominating the shared gradient update, even when directions are aligned (For instance, when one task contributes a gradient of 100 while the average across the others is only 0.1).

\textbf{Task-Adaptive
Sparsity Control Module.} To adapt to task-specific conflict levels, we propose this module. Taking the selection of conflicting parameters as an example, the procedure is as follows:

First, we flatten the conflict score matrix $C_{\text{conflict}}(T_i)$ for task $T_i$, and sort it in ascending order to obtain a vector as $\mathbf{X}^{T_i}$. Let $n$ be the total number of parameters. For each task $T_i$, we compute the quantile $Q^{T_i}_q$ as:
\begin{equation}
Q^{T_i}_q =
\begin{cases}
\mathbf{X}^{T_i}_{(k)}, & \text{if } k = qn \in \mathbb{Z} \\
(1 - \gamma) \cdot \mathbf{X}^{T_i}_{(\lfloor k \rfloor)} + \gamma \cdot \mathbf{X}^{T_i}_{(\lceil k \rceil)}, & \text{if } k = qn \notin \mathbb{Z} 
\end{cases}
\end{equation}
where $\gamma = qn - \lfloor qn \rfloor$ , $ k = qn$ and $\mathbf{X}_{(i)}$ denotes the $i$-th smallest element in $\mathbf{X}$. Using the quantiles, we define the interquartile range as:

\begin{equation}
\label{Thres}
\operatorname{IQR}^{T_i} = Q^{T_i}_{q3} - Q^{T_i}_{q1},  
\end{equation}
Second, a dynamic threshold is computed as:
\begin{equation}
\label{Thre}
\text{Threshold}_{T_i} = Q^{T_i}_{q1} - \beta_t \times \operatorname{IQR}^{T_i}.
\end{equation}
here, \( \beta_t \) is a dynamic coefficient that evolves during training, and \( q_1 \), \( q_3 \) are predefined hyperparameters. Parameters with scores below the threshold are considered significantly conflicting:
\begin{equation}
\label{Ci}
\mathcal{C}_i = \big\{ j \mid C_{\text{conflict}}(T_i)_j < \text{Threshold}_{T_i} \big\}.
\end{equation}

To improve the stability and efficiency of conflict-aware mask updates, we first introduce a general dynamic modulation function based on \textit{cosine annealing}~\cite{loshchilov2016sgdr}, defined as:
\begin{equation}
g(\eta_{\max}, \eta_{\min}) = 
\eta_{\min} + \frac{1}{2}(\eta_{\max} - \eta_{\min}) 
\left[ 1 + \cos\left( 2\pi \frac{t}{T} \right) \right],
\end{equation}
where $g(\cdot)$ denotes the standard cosine annealing function, $t$ is the current iteration step, $T$ is the total number of iterations, and $\eta_{\max}, \eta_{\min}$ are the upper and lower bounds of the annealing schedule, respectively.

To account for the asymmetric tolerance toward conflicting parameters at different training phases, we propose a \textit{piecewise asymmetric cosine annealing strategy} to dynamically generate the update coefficient $\beta_t$. Specifically:
\begin{equation}
\label{beta}
\beta_t =
\begin{cases}
g(\beta_{\text{left\_max}}, \beta_{\min}, t, T) & t \leq T/2, \\[6pt]
g(\beta_{\text{right\_max}}, \beta_{\min}, t, T) & t > T/2,
\end{cases}
\end{equation}
where $\beta_{\text{left\_max}} > \beta_{\text{right\_max}}$ denote the maximum conflict tolerance in the early and late stages, respectively, and $\beta_{\min}$ is the minimum threshold.
This design adaptively adjusts the masking sparsity throughout training. In the early stage, a larger $\beta_t$ (from $\beta_{\text{left\_max}}$) prevents excessive pruning under noisy conflict estimates; as training stabilizes, a smaller $\beta_t$ promotes fine-grained conflict masking. Toward the end, $\beta_t$ slightly increases to ensure smooth convergence.

Finally, we apply the same TAMU to the harmony score $C_{\text{harmony}}(T_i)$ of each task $T_i$. Specifically, we compute $Q^{T_i}_{q1}$, $Q^{T_i}_{q3}$ and $\operatorname{IQR}^{T_i}$ of the harmony scores, and define a task-specific harmony threshold as:
$\text{Threshold}_{T_i} = Q^{T_i}_{q3} + \beta_t \times \operatorname{IQR}^{T_i}$.
Parameters with scores exceeding this threshold are considered recoverable harmonious parameters and are collected into the set $\mathcal{R}_i$.

\textbf{Mask Update.} After identifying the conflicting and harmonious parameters, we proceed to update the mask matrix for each task. For each newly identified conflicting parameter $\mathcal{C}_i$, we adopt a soft masking strategy by setting its mask value to the normalized Fisher information, as follows:

\begin{equation}
\label{Mmask}
M^{T_i}_{\text{mask}, j} = 
\begin{cases}
\frac{F(T_i)_j - F(T_i)_{\min}}{F(T_i)_{\max} - F(T_i)_{\min}} & \text{if } j \in \mathcal{C}_i. \\
0 & \text{otherwise}.
\end{cases}
\end{equation}

Meanwhile, for parameters in the recovered harmonious set $\mathcal{R}_i$, we assign their mask values in the following form to facilitate consistent and unified mask updates:

\begin{equation}
\label{Mrecover}
M^{T_i}_{\text{recover}, j} =
\begin{cases}
1 - M^{T_i}_{j} & \text{if } j \in \mathcal{R}_i, \\
0 & \text{otherwise}.
\end{cases}
\end{equation}

Finally, we perform the mask update via matrix addition and subtraction. Specifically, we replace the original mask values of conflicting parameters $\mathcal{C}_i$ with their newly computed soft values, and restore the recovered harmonious parameters $\mathcal{R}_i$ by setting their mask values to 1. The update rule is formulated as:
\begin{equation}
M^{T_i} = M^{T_i} - (M^{T_i}_{\mathcal{C}_i} - M^{T_i}_{\text{mask}}) + M^{T_i}_{\text{recover}}.
\end{equation}

\begin{algorithm}[htbp]
\caption{SoCo-DT Training Framework}
\label{alg:erk-finegrained-mask}
\textbf{Input}: Number of tasks $N$, max training epochs $E$, mask update interval $interval_{mask}$, adaptive sparsity hyperparameters $\beta_{\text{left\_max}}, \beta_{\text{right\_max}}, \beta_{\min}$\\
\textbf{Output}: Task-specific optimized masks $M^{T_i}$ and shared parameters $\theta$

\begin{algorithmic}[H]
\small
\STATE Initialize model parameters $\theta$ and mask matrices $M$
\FOR{each step $t = 1$ to $E$}
    \IF{$t \bmod interval_{mask} == 0$}
        \STATE Compute average gradient:\\ $\bar{\mathbf{g}} = \frac{1}{N} \sum_{i=1}^N \nabla \mathcal{L}_{T_i}(\theta \odot \widetilde{M}^{T_i})$
        \FOR{each task $T_i$}
            \STATE Compute task gradient: $\mathbf{g}_i = \nabla \mathcal{L}_{T_i}(\theta \odot \widetilde{M}^{T_i})$
            \STATE Compute Fisher importance $F(T_i)$ use Eq.~\eqref{fisher}
            \STATE Compute conflict score $C_{\text{mask}}$ and harmony score $C_{\text{recover}}$ use Eq.~\eqref{Cconflict} and Eq.~\eqref{Charmony}
            \STATE Update $\beta_t$ by Eq.~\eqref{beta}
            \STATE Determine $\text{Threshold}_{T_i}$ use Eq.~\eqref{Thres}
            \STATE Identify conflict indices $\mathcal{C}_i$ use Eq.~\eqref{Ci}
            \STATE Similarly, compute harmony indices $\mathcal{R}_i$ for recovery
            \STATE Assign $M^{T_i}_{\text{mask}}$ and $M^{T_i}_{\text{recover}}$ use Eq.~\eqref{Mmask} and Eq.~\eqref{Mrecover}
            \STATE Final mask update:\\ $M^{T_i} = M^{T_i} - (M^{T_i}_{\mathcal{C}_i} - M^{T_i}_{\text{mask}}) + M^{T_i}_{\text{recover}}$
        \ENDFOR
    \ENDIF
    \STATE \texttt{// Parameter Update with Masking}
    \STATE $\theta_{t+1} = \theta_t - \eta \, \mathbb{E}\left[ \nabla \mathcal{L}_{\mathcal{T}_i}(\theta \odot \widetilde{M}^{T_i}) \right] \odot M^{T_i}$
\ENDFOR
\end{algorithmic}
\end{algorithm}

\begin{table*}[t]
\centering
\caption{Performance comparison on Meta-World with 5, 30, and 50 randomly sampled tasks under near-optimal and sub-optimal datasets. Each result is averaged over three random seeds with 50 evaluations per task. To ensure fair comparison, we adopt the same random seeds as those used in the previous baseline methods.}
\begin{tabular}{l|cc|cc|cc}
\toprule[2pt]
Method & \multicolumn{2}{c|}{Meta-World 5 Tasks} & \multicolumn{2}{c|}{Meta-World 30 Tasks} & \multicolumn{2}{c}{Meta-World 50 Tasks} \\
\cmidrule{2-7}
& Near-optimal & Sub-optimal & Near-optimal & Sub-optimal & Near-optimal & Sub-optimal \\
\midrule
MTDIFF  & $\textbf{100.0} \pm 0.0$ & $66.30 \pm 2.31$ & $67.52 \pm 0.35$ & $54.21 \pm 1.10$ & $61.32 \pm 0.89$ & $48.94 \pm 0.95$ \\
MTDT & $\textbf{100.0} \pm 0.0$ & $64.67 \pm 5.25$ & $71.89 \pm 0.95$ & $49.33 \pm 2.05$ & $65.80 \pm 1.02$ & $42.33 \pm 1.89$ \\
Prompt-DT*  & $\textbf{100.0} \pm 0.0$ & $67.00 \pm 2.33$ & $71.30 \pm 0.67$ & $54.33 \pm 0.78$ & $71.60 \pm 1.40$ & $51.40 \pm 0.35$ \\
HarmoDT-R*  & $\textbf{100.0} \pm 0.0$ & $68.00 \pm 2.33$ & $80.10 \pm 3.12$ & $59.00 \pm 3.66$ & $74.24 \pm 2.66$ & $52.45 \pm 1.17$ \\
HarmoDT-M*  & $\textbf{100.0} \pm 0.0$ & $72.12 \pm 3.48$ & $79.67 \pm 1.46$ & $\textbf{62.33} \pm 0.66$ & $\textbf{78.80} \pm 0.67$ & $56.20 \pm 0.56$ \\
HarmoDT-F*  & $\textbf{100.0} \pm 0.0$ & $\textbf{75.04} \pm 2.33$ & $\textbf{81.60} \pm 0.78$ & $60.60 \pm 1.12$ & $77.80 \pm 0.66$ & $\textbf{57.50} \pm 0.48$ \\
\midrule
\textbf{SoCo-DT (Ours)}  & $\textbf{100.0} \pm 0.0$ & $\textbf{80.00} \pm 2.32$ & $\textbf{87.38} \pm 1.20$ & $\textbf{69.60} \pm 1.12$ & $\textbf{85.88} \pm 0.86$ & $\textbf{67.19} \pm 0.90$ \\
\bottomrule[2pt]
\end{tabular}

\label{tb:scale}
\end{table*}
\section{Experiments}
\label{sec:experiments}

In this section, we conduct comprehensive experiments on the Meta-World~\cite{yu2020meta} benchmark (covering MT5, MT30, and MT50 configurations) to answer the following key research questions:
(1) Can the proposed SoCo-DT method outperform other mainstream multi-task offline RL algorithms on both near-optimal and sub-optimal datasets?
(2) How does the diversity of random task combinations affect the stability and generalization capability of our model under MT5 and MT30 settings?
(3) How do different design choices and components, such as the soft mask mechanism, task-adaptive sparsity control module and magnitude-sensitive harmony scoring mechanism, influence overall performance?

\subsection{Environment and Baselines}
We evaluate our method on the Meta-World multi-task robotic manipulation benchmark, which includes 50 tasks with shared dynamics and diverse object interactions. Following recent works~\cite{hu2024harmodt}~\cite{he2023diffusion}, we adopt random goal settings to assess task generalization. The evaluation metric is the average success rate across tasks.

We consider two types of offline datasets for training: (1) \textbf{Near-optimal dataset}: generated by a SAC-Replay~\cite{haarnoja2018soft} policy that mixes random and expert trajectories, resulting in high-quality data overall. (2) \textbf{Sub-optimal dataset}: composed of early-stage trajectories with only 50\% expert demonstrations retained, representing a more challenging and realistic scenario.

\textbf{Baselines.} We compare our method against several representative multi-task offline RL baselines:
(1) \textbf{MTBC}~\cite{he2023diffusion}: Multi-task Behavior Cloning with task ID conditioning.
(2) \textbf{MTIQL}: Multi-task IQL~\cite{kostrikov2021offline} with multi-head critic and task-conditioned policy network.
(3) \textbf{MTDIFF-P}~\cite{he2023diffusion}: A diffusion-based policy learning method with prompting and transformer modules.
(4) \textbf{MTDT}~\cite{he2023diffusion}: Decision Transformer adapted for multi-task learning with task ID embedding.
(5) \textbf{Prompt-DT}~\cite{xu2022prompting}: Decision Transformer with trajectory prompt and reward-to-go for unseen tasks.
Additionally, we include four representative online RL methods for reference:
(6) \textbf{CARE}~\cite{sodhani2021multi}: Uses meta-data and encoder mixing for task representation.
(7) \textbf{PaCo}~\cite{sun2022paco}: Adopts parameter composition for task-specific parameter reassembly.
(8) \textbf{Soft-M}~\cite{yang2020multi}: Uses routing networks for modular soft composition.
(9) \textbf{D2R}~\cite{he2024not}: Employs diverse routing paths for different tasks.(10) \textbf{HarmoDT}~\cite{hu2024harmodt}: as discussed in \S~\ref{harmodt}. We use results from ~\cite{he2023diffusion} for most baselines, * indicate baselines of our own implementation.

\subsection{Performance Comparison}
We report performance comparisons with mainstream methods on MT5, MT30, and MT50. As shown in Table~\ref{tb:scale}, our method consistently outperforms all baselines across different task scales. Specifically, it achieves 100\% success rate on the near-optimal MT5 dataset and gains an 
improvement of 4.96\% on the sub-optimal dataset. For MT30, our method shows gains of 5.78\% and 7.27\% on the near-optimal and sub-optimal datasets, respectively. For MT50, it achieves 7.08\% and 9.69\% improvements. Notably, the performance gains are larger on sub-optimal datasets, indicating stronger stability and robustness when reward signals are ambiguous.
\begin{table}[h]
\centering
\caption{Average success rate across 3 seeds on MT50 with random goals (MT50-rand) under both near-optimal and sub-optimal cases.
Each task is evaluated for 50 episodes.}
\vspace{0.1cm}
\small
\begin{tabular}{l|cc}
\toprule[2pt]
Method & Near-optimal & Sub-optimal \\
\midrule 
\textbf{CARE}~(online) & $46.12_{\pm 1.30}$ & -- \\
\textbf{PaCo}~(online) & $54.31_{\pm 1.32}$ & -- \\
\textbf{Soft-M}~(online) & $53.41_{\pm 0.72}$ & -- \\
\textbf{D2R}~(online) & $63.53_{\pm 1.22}$ & -- \\
\midrule
\textbf{MTBC}  & $60.39_{\pm 0.86}$ & $34.53_{\pm 1.25}$ \\
\textbf{MTIQL} & $56.21_{\pm 1.39}$ & $43.28_{\pm 0.90}$ \\
\textbf{MTDIFF-P} & $59.53_{\pm 1.12}$ & $48.67_{\pm 1.32}$ \\
\textbf{MTDIFF-P-ONEHOT} & $61.32_{\pm 0.89}$ & $48.94_{\pm 0.95}$ \\
\textbf{MTDT} & $65.80_{\pm 1.02}$ & $42.33_{\pm 1.89}$ \\
\textbf{Prompt-DT*} & $71.60_{\pm 1.40}$ & $51.40_{\pm 0.35}$ \\
\textbf{HarmoDT-R*} & $74.24_{\pm 2.66}$ & $52.45_{\pm 1.17}$ \\
\textbf{HarmoDT-M*} & $\textbf{78.80}_{\pm 0.67}$ & $56.20_{\pm 0.56}$ \\
\textbf{HarmoDT-F*} & $77.80_{\pm 0.66}$ & $\textbf{57.50}_{\pm 0.48}$ \\
\midrule
\textbf{SoCo-DT(ours)} & $\textbf{85.88}_{\pm 0.86}$ & $\textbf{67.19}_{\pm 0.90}$ \\
\bottomrule[2pt]
\end{tabular}
\label{tb:50task}
\end{table}
\begin{figure}[t]
  \centering
 \begin{subfigure}{0.23\textwidth}
  \centering
  \includegraphics[width=\linewidth]{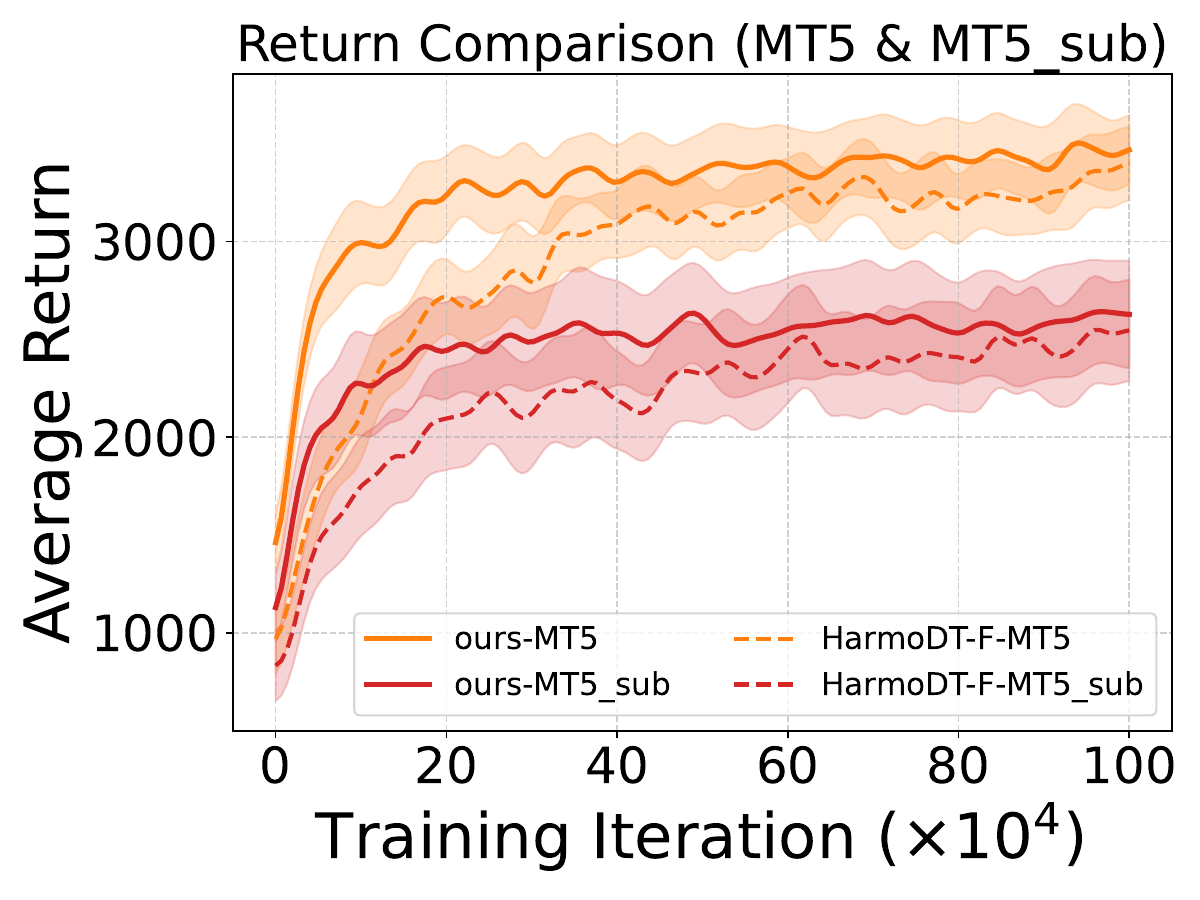}
  \caption{Return comparison on MT5}
  \label{fig:mt5_return}
\end{subfigure}
\hfill
\begin{subfigure}{0.23\textwidth}
  \centering
  \includegraphics[width=\linewidth]{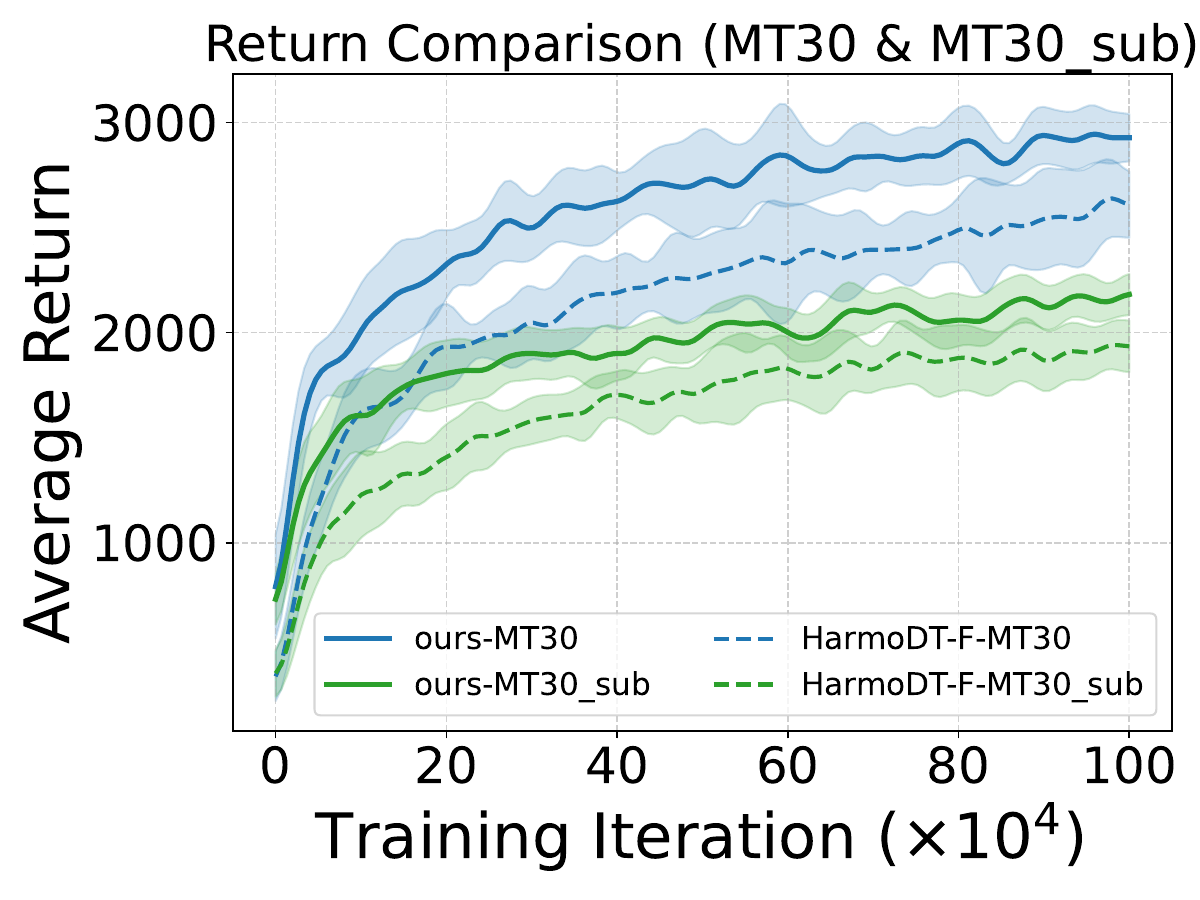}
  \caption{Return comparison on MT30}
  \label{fig:mt30_return}
\end{subfigure}
  \caption{Average return comparison for MT5 and MT30 task settings. (a) shows average return curves for MT5 and its sub-task variant; (b) shows the same for MT30. Both are compared with the state-of-the-art method HarmoDT-F.}
  \label{fig:masked_combined}
\end{figure}

\begin{table}[h]
\small
\centering
\caption{Performance comparison on random task combinations. Results are averaged over three seeds.}
\label{tab:meta_world_vertical_compact}
\begin{tabular}{l|l|c}
\toprule[2pt]
\textbf{Setting} & \textbf{Method} & \textbf{Success Rate} \\
\midrule
\multirow{2}{*}{5 Tasks} 
  & HarmoDT-F & 
    \begin{tabular}[c]{@{}c@{}}Near-opt: $95.33 \pm 5.25$ \\ Sub-opt: $74.67 \pm 7.37$\end{tabular} \\
\cmidrule(lr){2-3}
  & \textbf{SoCo-DT (Ours)} & 
    \begin{tabular}[c]{@{}c@{}}Near-opt: $\textbf{96.20} \pm 4.42$ \\ Sub-opt: $\textbf{79.43} \pm 13.77$\end{tabular} \\
\midrule
\multirow{2}{*}{30 Tasks} 
  & HarmoDT-F & 
    \begin{tabular}[c]{@{}c@{}}Near-opt: $81.30 \pm 3.12$ \\ Sub-opt: $63.70 \pm 2.35$\end{tabular} \\
\cmidrule(lr){2-3}
  & \textbf{SoCo-DT (Ours)} & 
    \begin{tabular}[c]{@{}c@{}}Near-opt: $\textbf{86.27} \pm 2.58$ \\ Sub-opt: $\textbf{67.63} \pm 1.10$\end{tabular} \\
\bottomrule[2pt]
\end{tabular}
\end{table}
To further verify robustness under varying task combinations, we randomly sample 5 and 30 tasks from the Meta-World using three random seeds and compare with HarmoDT-F. Results are shown in Table~\ref{tab:meta_world_vertical_compact} and Figure~\ref{fig:masked_combined}.

\subsection{Ablation Study}
To thoroughly evaluate the contributions of each module and hyperparameter in our method, we conduct ablation studies on the near-optimal dataset under the MT30 task configuration. The ablations cover both architectural components and key hyperparameters. We report the results of method-level and parameter-level ablations separately below.

\textbf{Component Ablation.} We evaluate the average success rate of model variants by ablating or replacing key components under the same environment:
\textbf{(1) Without Soft Mask Mechanism}: The soft mask is replaced with a hard mask, where the conflicting parameters are directly set to zero without considering their importance weights.\textbf{(2) Fixed Mask Sparsity Mechanism}: The mask sparsity is fixed to a predefined value, instead of being dynamically adjusted using the IQR-based thresholding. A fixed proportion of parameters is masked in each iteration. \textbf{(3) Without Magnitude-Sensitive Harmony Scoring Mechanism}: During the recovery phase, parameter scoring is performed using the conventional dot product method, without employing the magnitude-sensitive harmony scoring mechanism. The experimental results are presented in Table~\ref{tab:ablation_results1}.

\begin{table}[h]
\centering
\caption{Results of the component ablation study.  
\textbf{S}, \textbf{A}, and \textbf{M} denote the \textit{Soft Mask Mechanism}, \textit{Adaptive Mask Sparsity Mechanism}, and \textit{Magnitude-Sensitive Harmony Scoring Mechanism}, respectively. The table reports average success rate and return for different component combinations on both near-optimal and sub-optimal datasets under the MT30.}
\label{tab:ablation_results1}
\resizebox{0.98\linewidth}{!}{
\begin{tabular}{l|l|c|c}
\toprule[2pt]
\textbf{Datasets} & \textbf{Method} & \textbf{Avg. Success Rate} & \textbf{Avg. Return} \\
\midrule
\multirow{4}{*}{near-optimal} 
 & A+M   & 72.00 & 2549.8\\
 & S+M  & 76.33 & 2635.7 \\
 & S+A   & 76.00 & 2577.8 \\
 & \textbf{S+A+M} & \textbf{86.33} & \textbf{3072.7} \\
\midrule
\multirow{4}{*}{sub-optimal} 
 & A+M   & 61.33 &  2094.5\\
 & S+M   & 59.67 & 2037.8 \\
 & S+A   & 59.33 & 1871.1 \\
 & \textbf{S+A+M} & \textbf{69.60} & \textbf{2417.7} \\
\bottomrule[2pt]
\end{tabular}
}
\end{table}
\begin{figure}[h]
  \centering
  \begin{subfigure}{0.48\linewidth}
    \centering
    \includegraphics[width=\linewidth]{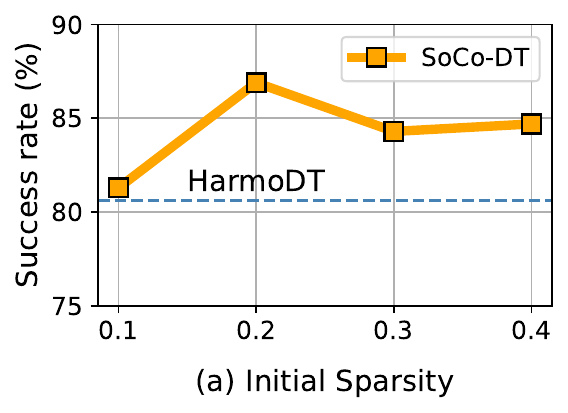}
  \end{subfigure}
  \hfill
  \begin{subfigure}{0.48\linewidth}
    \centering
    \includegraphics[width=\linewidth]{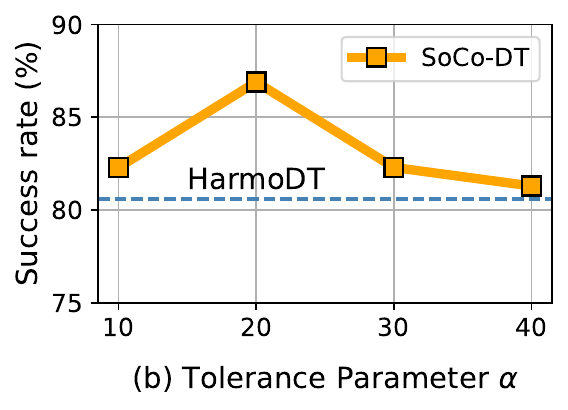}
  \end{subfigure}
  \\
  \begin{subfigure}{0.48\linewidth}
    \centering
    \includegraphics[width=\linewidth]{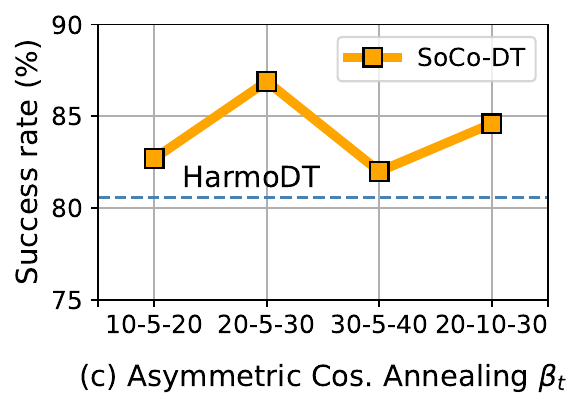}
  \end{subfigure}
  \hfill
  \begin{subfigure}{0.48\linewidth}
    \centering
    \includegraphics[width=\linewidth]{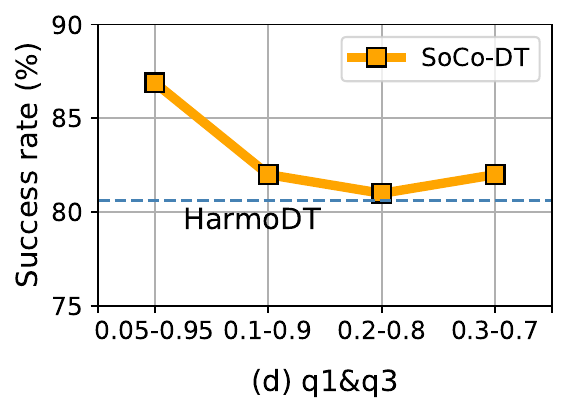}
  \end{subfigure}

  \caption{
    Ablation study under the near-optimal setting on the MT30.
    Default values: initial sparsity = 0.2, $\alpha = 20$, $\beta_{\text{left\_max}}$ = 20, $\beta_{\min}$ = 5, $\beta_{\text{right\_max}}$ = 30, $q_1$ = 0.05,
    $q_3$ = 0.95.
  }
  \label{fig:XR}
\end{figure}

\textbf{Hyperparameter Ablation.} We further analyze the impact of key hyperparameters on the model’s performance. The hyperparameters include:
\textbf{(1) Initial Mask Sparsity}: Specifies the initial proportion of parameters to be masked, which controls the starting level of sparsity in the model.
\textbf{(2) Tolerance Parameter $\alpha$}: Determines the level of tolerance for gradient magnitude discrepancy in the same direction during parameter recovery, see Eq.~\eqref{Hti}.
\textbf{(3) Asymmetric Cosine Annealing for $\beta_t$}: Controls the dynamic adjustment range of the conflict and harmony thresholds during the selection of conflicting and harmonious parameters, see Eq.~\eqref{Thre}.
\textbf{(4) Initial Quantile Values $q_1$ and $q_3$}: Define the initial quantile points used in the IQR method for task-adaptive threshold computation, see Eq.~\eqref{Thres}. The experimental results are illustrated in Figure~\ref{fig:XR}.

\section{Conclusion}
In this study, we identify limitations in existing mask-based methods for multi-task reinforcement learning and propose a soft masking mechanism to reduce the negative impact of traditional hard masks on task-critical parameters, improving model generalization and learning efficiency. We also introduce a task-aware mask update with adaptive sparsity strategy, dynamically adjusting mask sparsity based on task complexity. Extensive experiments demonstrate the superior
performance of our proposed method.

\section{Acknowledgments}
This work was supported by the National Key Research and Development Program of China (Grant Nos. 2021YFA1000102 and 2021YFA1000103), the Shandong Provincial Natural Science Foundation (Grant Nos. ZR2024MF129 and ZR2025QC1540), and the Fundamental Research Funds for the Central Universities (Grant No. 25CX06034A).

\bibliography{aaai2026}
\end{document}